\documentclass[10pt,twocolumn]{article}

\usepackage[letterpaper,margin=0.72in]{geometry}
\usepackage[T1]{fontenc}
\usepackage[utf8]{inputenc}
\usepackage{newtxtext,newtxmath}
\usepackage[hyphens]{url}
\usepackage{graphicx}
\usepackage{natbib}
\usepackage{caption}
\usepackage{authblk}
\usepackage{algorithm}
\usepackage{algorithmic}
\usepackage{newfloat}
\usepackage{listings}
\usepackage{booktabs}
\usepackage{amsmath}
\usepackage{tikz}

\usetikzlibrary{positioning,calc,arrows.meta}
\definecolor{cSin}{RGB}{53,126,199}
\definecolor{cCos}{RGB}{224,90,90}
\definecolor{cHist}{RGB}{53,126,199}
\definecolor{cPred}{RGB}{210,75,75}
\definecolor{cBlk}{RGB}{245,245,245}

\DeclareCaptionStyle{ruled}{labelfont=normalfont,labelsep=colon,strut=off}
\floatstyle{ruled}
\newfloat{listing}{tb}{lst}{}
\floatname{listing}{Listing}

\title{CAMP: A Cycle-Aware Multi-Scale Patch Mixer for Time Series Forecasting}

\author{Jung Min Choi\thanks{Corresponding author: \texttt{choi@ismll.de}}}
\author{Vijaya Krishna Yalavarthi}
\author{Lars Schmidt-Thieme}
\affil{ISMLL, University of Hildesheim, VWFS Data Analytics Research Center (VWFS-DARC), Hildesheim, Germany\\
\texttt{choi@ismll.de}, \texttt{yalavarthi@ismll.de}, \texttt{schmidt-thieme@ismll.de}}
\date{}

\begin{document}
\maketitle

\begin{abstract}
    Real-world time series are often governed by recurring patterns, but their dominant periods may vary across datasets, forecasting settings, and individual input windows. Existing cycle-aware forecasters commonly rely on a single period selected at the dataset level, which can be restrictive when periodic behavior changes over time or when multiple cycles coexist. 
    Moreover, patch-based models typically process all patch positions uniformly, although patches farther from the forecast boundary may require broader contextual refinement, while recent patches contain information that should be preserved more directly. 
    After cyclic behavior is removed, the remaining dynamics may also span multiple temporal resolutions and cannot be adequately described at a single scale. 
    We introduce CAMP, a \textbf{C}ycle-\textbf{A}ware \textbf{M}ulti-Scale \textbf{P}atch Mixer designed to address these challenges. The Adaptive Cycle Learning module identifies dominant frequencies separately for each input window and generates both historical and future cyclic components without requiring a predefined cycle length. The Horizon-Guided Patch Mixer introduces position-dependent refinement, allowing earlier patches to incorporate broader temporal context while preserving information close to the forecast boundary. CAMP further models the de-cycled residual through temporally aligned multi-resolution representations, enabling complementary dynamics at different scales to be captured within one forecasting framework. Across seven long-term forecasting benchmarks, CAMP achieves the best average MSE on six datasets and the best or tied-best MAE on six. It also obtains the highest MSE win count across sixteen settings on four PEMS traffic benchmarks. Ablation studies confirm the complementary contributions of sample-adaptive cycle learning, multi-resolution residual modeling, intra-patch encoding, and horizon-guided patch refinement.
    \end{abstract}

\section{Introduction}

\begin{figure}[t]
    \centering
    \includegraphics[width=0.82\linewidth]{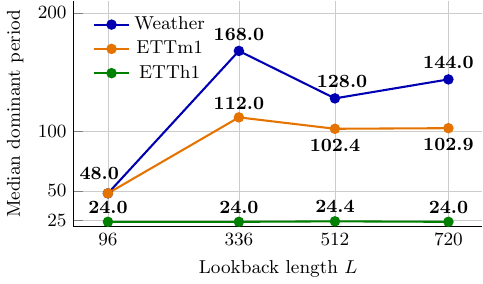}
    \caption{Median dominant period across test input windows for different
look-back lengths. For each window, the highest-amplitude non-DC FFT
frequency is converted to its corresponding period. The variation across
datasets and input lengths motivates adaptive cycle identification.}
    \label{fig:dominant_cycle_analysis}
\end{figure}

Time series forecasting (TSF) supports applications such as electricity consumption, weather forecasting, traffic monitoring, and industrial sensing~\cite{weather,electricity,pems,informer}.
Multivariate time series (MTS) forecasting is challenging because variables exhibit distinct but interacting dynamics, while real-world sequences combine trends, recurring cycles, and irregular residual fluctuations.
Failing to capture these structures can distort both future trajectories and recurring behavior.

Recent studies show that lightweight MLP architectures can model complex temporal dependencies without attention.
TSMixer~\cite{tsmixer,tsmixer2} learns temporal and cross-variable interactions, while TimeMixer~\cite{timemixer} models variations across multiple scales.
Other approaches exploit hierarchical patch dependencies in HDMixer~\cite{hdmixer}, wavelet-based representations in WPMixer~\cite{wpmixer}, and frequency-domain learning in FreTS~\cite{frets}.

Cyclic behavior is particularly useful because recurring patterns can be extrapolated into future horizons.
CycleNet~\cite{cyclenet} and FreqCycle~\cite{freqcycle} demonstrate the benefit of explicitly modeling periodic components.
However, they typically require cycle lengths to be predefined or estimated before training through dataset-level analyses such as the autocorrelation function (ACF).
This can be restrictive because dominant periodicities may vary across datasets, look-back lengths, and input windows, while multiple cycles may coexist.

Figure~\ref{fig:dominant_cycle_analysis} summarizes how the dominant period changes with the look-back length.
For each dataset and look-back length $L$, we compute the channel-averaged FFT of every test input window, select the highest-amplitude non-DC frequency bin $i$, convert it to the nominal period $L/i$, and report the median across all windows.
The median dominant period varies substantially for Weather and ETTm1 as the look-back length changes, whereas ETTh1 remains close to 24.
Thus, a single predefined cycle may not capture the relevant periodic structure across forecasting settings, motivating adaptive frequency identification for each input sequence.

We propose \textbf{CAMP}, a \textbf{C}ycle-\textbf{A}ware \textbf{M}ulti-Scale \textbf{P}atch Mixer that separately models cyclic and residual dynamics.
Its Adaptive Cycle Learning Module identifies sample-specific dominant frequencies using the Fast Fourier Transform (FFT)~\cite{fft} and generates historical and future cyclic components through an MLP.
CAMP therefore requires neither a predefined cycle length nor a separate dataset-level cycle-selection step.

After removing the historical cycle, CAMP applies the Stationary Wavelet Transform (SWT)~\cite{swt} to decompose the residual into aligned multi-resolution streams encoded by patch-level MLP mixers.
The proposed \textit{Horizon-Guided Patch Mixer} progressively refines earlier patches of the coarsest detail stream while preserving patches near the forecasting horizon.
The encoded streams are aggregated to predict the residual and combined with the future cyclic component.

The main contributions are:
\begin{itemize}
    \item We propose \textbf{CAMP}, a cycle-residual forecasting model that identifies sample-specific dominant frequencies and generates historical and future cycles without predefined cycle lengths.

    \item We combine SWT-based multi-resolution decomposition, patch-level MLP encoding, and horizon-guided patch mixing to model complementary residual dynamics.

    \item CAMP achieves strong performance on both long-term and short-term forecasting benchmarks.
\end{itemize}

\begin{figure*}[t]
    \centering
    \includegraphics[width=\textwidth]{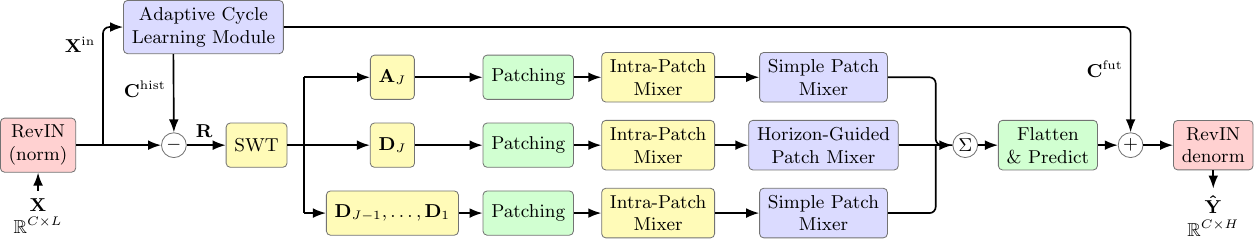}
    \caption{Overview of CAMP. The RevIN-normalized input is decomposed
        into a learned cyclic component and a residual. The residual is
        separated into aligned multi-resolution streams by SWT. The coarsest
        detail stream is encoded by the Horizon-Guided Patch Mixer, whereas
        the remaining streams use Simple Patch Mixers. Their representations
        are averaged and mapped to the residual forecast, after which the
        predicted cyclic component is reinjected before RevIN
        denormalization.}
    \label{fig:camp}
\end{figure*}
\section{Related Work}

\paragraph{MLP-Based Time Series Forecasting Models.}
MLP-based models achieve competitive forecasting performance without complex attention mechanisms.
TSMixer~\cite{tsmixer,tsmixer2} learns temporal and channel dependencies through MLP mixing, while TimeMixer~\cite{timemixer} captures variations across decomposed temporal scales.
Recent models further combine MLP mixing with structured representations: HDMixer~\cite{hdmixer} models hierarchical dependencies through extendable patching, and WPMixer~\cite{wpmixer} integrates wavelet decomposition with patch-level mixing.
CAMP follows this direction but explicitly separates dominant cyclic behavior from residual dynamics.

\paragraph{Cycle and Periodicity Learning.}
Cycle-aware forecasting models exploit recurring temporal patterns in
different ways. TimesNet~\cite{timesnet} identifies dominant spectral
periods and constructs period-based representations, while
DEPTS~\cite{depts} learns parameterized periodic and residual components.
SparseTSF~\cite{sparsetsf} decouples periodicity and trend through
cross-period sparse forecasting. Koopa~\cite{koopa} separates
time-invariant and time-varying dynamics using a Fourier filter and
Koopman predictors, while BasisFormer~\cite{basisformer} learns
interpretable historical and future bases. CycleNet~\cite{cyclenet} and
FreqCycle~\cite{freqcycle} explicitly model recurrent cycles but require
predefined or pre-estimated periods. CAMP instead selects sample-specific
dominant frequencies and directly generates historical and future cyclic
components without fixing a global cycle length.

\paragraph{Patching in Time Series Forecasting.}
The use of patches as tokens was popularized by Vision Transformer~\cite{vit} and later adapted to time series forecasting.
Patching groups adjacent observations into local temporal segments, reducing sequence length while preserving local structure.
PatchTST~\cite{patchtst} learns patch representations instead of processing individual time steps, while SRSNet~\cite{srsnet} introduces selective patching and dynamic reassembly to retain informative segments.
CAMP applies patch based MLP encoders to residual streams and introduces a Horizon Guided Patch Mixer that preserves recent patches while progressively refining earlier information.

\paragraph{Frequency-Domain Forecasting.}
Frequency-domain representations expose periodic and global structures
that may be less apparent in the time domain. FEDformer~\cite{fedformer}
combines temporal decomposition with frequency-enhanced modeling, while
FiLM~\cite{film} uses Fourier projection to preserve informative
historical patterns. FreTS~\cite{frets} applies MLPs to frequency-domain
representations, and FourierGNN~\cite{fouriergnn} models multivariate
dependencies through Fourier graph operators. Recent approaches such as
TimeKAN~\cite{timekan}, Amplifier~\cite{amplifier},
WaveForM~\cite{waveform}, and SimpleTM~\cite{simpletm} further explore
frequency-specific and multi-resolution representations. CAMP differs by
applying SWT after adaptive cycle removal and encoding the aligned
residual streams through patch-level MLP mixers.

\section{Methodology}
\label{sec:methodology}

CAMP follows the cycle--residual formulation of
CycleNet~\cite{cyclenet}: a cyclic component is modeled explicitly,
removed from the input, and reinjected into the forecast, while a
patch-based backbone predicts the residual.
Unlike methods that use one cycle length for an entire dataset, CAMP
selects dominant frequencies independently for each input window.

This design has two main components.
The \textbf{Adaptive Cycle Learning} module selects per-sample frequency
bins using the Fast Fourier Transform (FFT) and synthesizes cyclic
representations over the look-back and prediction intervals.
The \textbf{Horizon-Guided Patch Mixer} encodes the residual using a
progressive schedule that applies fewer cross-patch updates near the
forecast boundary.
These components are combined with stationary wavelet decomposition and
stream-specific patch encoders.

\paragraph{Notation and Problem Formulation.}
Let $\mathbf{X}\in\mathbb{R}^{C\times L}$ denote a look-back window with
$C$ variables and $L$ time steps, and let
$\mathbf{Y}\in\mathbb{R}^{C\times H}$ denote the target over a prediction
horizon of length $H$.
Multivariate forecasting seeks
\begin{equation}
    F:\mathbb{R}^{C\times L}\rightarrow\mathbb{R}^{C\times H},
    \qquad
    \hat{\mathbf{Y}}=F(\mathbf{X}).
    \label{eq:problem}
\end{equation}
The batch dimension is omitted.
In implementation, the model receives $[B,L,C]$ inputs and returns
$[B,H,C]$ forecasts.
RevIN~\cite{revin} first normalizes the input to obtain
$\mathbf{X}^{\mathrm{in}}\in\mathbb{R}^{C\times L}$, and its inverse is
applied to the final prediction.

\paragraph{Patching.}
Given $\mathbf{Z}\in\mathbb{R}^{C\times L}$, patch length $P$, and stride
$S$, patching extracts $N$ potentially overlapping segments.
Replication padding completes the final segment when necessary, producing
$\mathbf{Z}^{p}\in\mathbb{R}^{C\times N\times P}$.

Each wavelet stream has an independent linear patch projection followed
by an Intra-Patch Mixer:
\begin{equation}
\begin{aligned}
    \mathbf{U}^{(0)}
    &=
    \mathrm{Linear}_{P\rightarrow D}
    \bigl(\mathbf{Z}^{p}\bigr),\\
    \mathbf{U}
    &=
    \mathbf{U}^{(0)}
    +
    \mathrm{MLP}_{\mathrm{intra}}
    \bigl(\mathbf{U}^{(0)}\bigr).
\end{aligned}
\label{eq:intra_patch}
\end{equation}
Here, $\mathbf{U}\in\mathbb{R}^{C\times N\times D}$.
The Intra-Patch Mixer acts only on the embedding dimension $D$ and does
not mix patch positions or variables.
All stream encoders use the same operator design but have independent
projection and mixer parameters.

We denote the $n$-th patch representation by
$\mathbf{U}_n\in\mathbb{R}^{C\times D}$ and a patch prefix by
$\mathbf{U}_{1:i}=[\mathbf{U}_1,\ldots,\mathbf{U}_i]$.
Patches are temporally ordered, with $\mathbf{U}_N$ closest to the
forecast boundary.

\begin{figure}[t]
    \centering
    \includegraphics[width=0.9\linewidth]{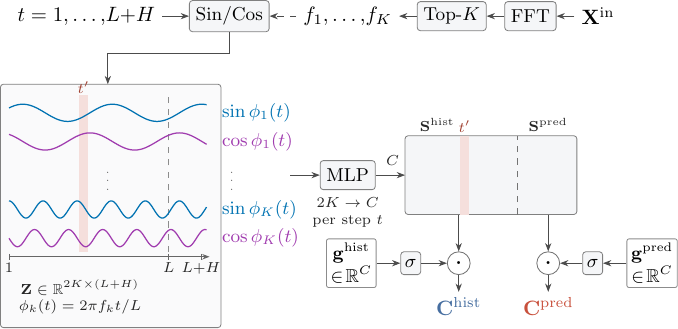}
    \caption{Architecture of Adaptive Cycle Learning. Per-sample
        top-$K$ FFT bins define sinusoidal features over the look-back and
        prediction intervals. A shared two-layer MLP maps the resulting
        $2K$ Fourier features to channel-wise cycle values. Separate sigmoid
        gates produce the historical cycle $\mathbf{C}^{\mathrm{hist}}$ and
        future cycle $\mathbf{C}^{\mathrm{pred}}$.}
    \label{fig:acl}
\end{figure}
\subsection{Adaptive Cycle Learning}
\label{sec:acl}

Dominant periodicities can vary across input windows, making a single
dataset level cycle insufficient. Adaptive Cycle Learning (ACL) instead
selects dominant frequency bins for each sample and uses them to condition
a shared cycle synthesis network. ACL assumes that the dominant frequency
support remains approximately stable from the look back window to the
prediction horizon, while allowing it to vary across windows.

\paragraph{Frequency Estimation.}
Given $\mathbf{X}^{\mathrm{in}}\in\mathbb{R}^{C\times L}$, we apply the
real-valued FFT along the temporal dimension and obtain the amplitude
spectrum $\boldsymbol{\Lambda}\in\mathbb{R}^{C\times F}$, where
$F=\lfloor L/2\rfloor+1$. Following the channel-averaged spectral
selection of TimesNet~\cite{timesnet}, but operating independently on
each sample, we compute
\begin{equation}
    \bar{\mathbf{a}}
    =
    \frac{1}{C}
    \sum_{c=1}^{C}
    \boldsymbol{\Lambda}_{c,:}
    \in \mathbb{R}^{F}.
    \label{eq:channel_spectrum}
\end{equation}
We suppress the DC component by setting $\bar{a}_0=0$ and select the
indices $(i_1,\ldots,i_K)$ of the $K$ largest entries of
$\bar{\mathbf{a}}$ in~\eqref{eq:channel_spectrum}. Bin $i_k$ corresponds
to normalized frequency $i_k/L$ and nominal period $L/i_k$.

Channel averaging emphasizes periodicities shared across variables and
therefore produces one harmonic basis for each sample. The discrete
Top-$K$ selection does not propagate gradients through the selected
indices. Learning instead occurs in the cycle synthesis network, the
global cycle scale, and the channel-wise gates.

\paragraph{Cycle Synthesis.}
For each selected bin $i_k$, we define the phase at time $t$ as
$\phi_k(t)=2\pi i_k t/L$. The corresponding sine and cosine values are
concatenated to form
\begin{equation}
    \mathbf{z}(t)
    =
    \operatorname{Concat}
    \left(
        [\sin \phi_k(t)]_{k=1}^{K},
        [\cos \phi_k(t)]_{k=1}^{K}
    \right)
    \in \mathbb{R}^{2K}.
    \label{eq:fourier_features}
\end{equation}
The Fourier feature vector in~\eqref{eq:fourier_features} is mapped to
channel-wise cycle values using a shared two-layer network:
\begin{equation}
    \mathbf{s}(t)
    =
    \gamma\,
    \mathrm{MLP}_{\mathrm{acl}}
    \bigl(\mathbf{z}(t)\bigr)
    \in \mathbb{R}^{C},
    \label{eq:cycle_synthesis}
\end{equation}
where $\mathrm{MLP}_{\mathrm{acl}}$ follows a
$2K\rightarrow D\rightarrow C$ architecture with GELU and dropout, and
$\gamma$ is a learnable global scale.

The sine and cosine pairs in~\eqref{eq:fourier_features} provide a
phase-complete harmonic basis, while the nonlinear mapping
in~\eqref{eq:cycle_synthesis} jointly combines the selected components.
ACL is sample-adaptive through the selected frequency bins, whereas the
synthesis network, global scale, and gates are shared across samples.
Because the network does not directly receive FFT amplitudes or complex
phases, ACL performs sample-specific frequency-support selection
followed by shared frequency-conditioned synthesis.

\paragraph{Extending to the Future Horizon.}
The same time parameterized function is evaluated over the historical
and future intervals:
\begin{equation}
\begin{aligned}
    \mathbf{S}^{\mathrm{hist}}
    &=
    [\mathbf{s}(0),\ldots,\mathbf{s}(L-1)],\\
    \mathbf{S}^{\mathrm{pred}}
    &=
    [\mathbf{s}(L),\ldots,\mathbf{s}(L+H-1)].
\end{aligned}
\label{eq:cycle_ranges}
\end{equation}
Accordingly,
$\mathbf{S}^{\mathrm{hist}}\in\mathbb{R}^{C\times L}$ and
$\mathbf{S}^{\mathrm{pred}}\in\mathbb{R}^{C\times H}$.
No horizon specific parameters are required.

CAMP applies separate channel wise gates to the historical and future
cycles:
\begin{equation}
\begin{aligned}
    \mathbf{C}^{\mathrm{hist}}
    &=
    \sigma(\mathbf{g}^{\mathrm{hist}})
    \odot
    \mathbf{S}^{\mathrm{hist}},\\
    \mathbf{C}^{\mathrm{pred}}
    &=
    \sigma(\mathbf{g}^{\mathrm{pred}})
    \odot
    \mathbf{S}^{\mathrm{pred}}.
\end{aligned}
\label{eq:cycle_gates}
\end{equation}
Here,
$\mathbf{g}^{\mathrm{hist}},\mathbf{g}^{\mathrm{pred}}
\in\mathbb{R}^{C}$,
$\sigma(\cdot)$ denotes the sigmoid function, and $\odot$ broadcasts the
gates over time. The gates are shared across samples but independently
learned for each variable and each side of the forecast boundary.

The initial global scale and gate logits are treated as hyperparameters
and selected using validation performance, while the final ACL layer is
initialized to zero. The generated cycle therefore starts at zero,
allowing CAMP to begin as a residual only model and gradually introduce
cyclic information during training.

\begin{figure}[t]
    \centering
    \includegraphics[width=0.8\linewidth]{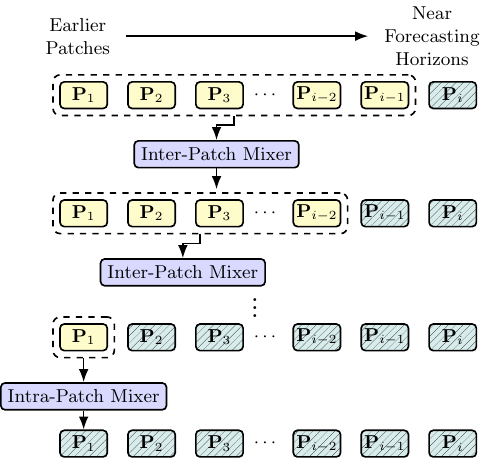}
    \caption{Architecture of the Horizon-Guided Patch Mixer.
    Earlier patches receive more progressive cross-patch updates, while
    patches closer to the forecasting horizon are updated less frequently.}
    \label{fig:progressive_patch_mixer}
\end{figure}
\subsection{Horizon-Guided Patch Mixer}
\label{sec:hgpm}

Standard patch mixers generally use the same update structure at every
position, although patches have different distances from the forecast
boundary.
The Horizon-Guided Patch Mixer (HGPM) instead preserves the most recent
patch while progressively mixing earlier context.

After patch projection and intra-patch refinement, HGPM processes
shrinking prefixes in the order $i=N-1,N-2,\ldots,1$.
At step $i$, only $\mathbf{U}_{1:i}$ is updated:
\begin{equation}
\begin{aligned}
    \mathbf{U}_{1:i}
    \leftarrow\;&
    \mathbf{U}_{1:i}
    +
    \boldsymbol{\alpha}_i
    \odot
    \Bigl(
    \mathcal{M}_i(\mathbf{U}_{1:i})
    -
    \mathbf{U}_{1:i}
    \Bigr).
\end{aligned}
\label{eq:hgpm}
\end{equation}
The learnable scale
$\boldsymbol{\alpha}_i\in\mathbb{R}^{C}$ is broadcast over patch and
feature dimensions.
It is implemented as an unconstrained channel-specific residual scale,
initialized to $0.01$ in all experiments and learned jointly with the
remaining model parameters.
Equation~\eqref{eq:hgpm} is therefore a residual correction, not
necessarily a convex interpolation.

For a prefix
$\mathbf{P}_i\in\mathbb{R}^{C\times i\times D}$,
the prefix mixer $\mathcal{M}_i$ performs feature, temporal, and channel
mixing sequentially. First, feature mixing is applied along the embedding
dimension $D$:
\begin{equation}
    \mathbf{U}_i
    =
    \mathbf{P}_i
    +
    \mathcal{F}_i(\mathbf{P}_i),
    \qquad
    \mathcal{F}_i:\mathbb{R}^{D}\rightarrow\mathbb{R}^{D}.
    \label{eq:hgpm_feature_mixing}
\end{equation}
Here, $\mathcal{F}_i$ is applied independently to every channel and patch
position.

Next, we reorder the tensor dimensions from
$\mathbb{R}^{C\times i\times D}$ to
$\mathbb{R}^{C\times D\times i}$ so that temporal mixing can be applied
along the $i$ patch positions:
\begin{equation}
    \mathbf{V}_i
    =
    \pi_t^{-1}
    \left(
        \pi_t(\mathbf{U}_i)
        +
        \mathcal{T}_i\bigl(\pi_t(\mathbf{U}_i)\bigr)
    \right),
    \qquad
    \mathcal{T}_i:\mathbb{R}^{i}\rightarrow\mathbb{R}^{i},
    \label{eq:hgpm_temporal_mixing}
\end{equation}
where
$\pi_t:\mathbb{R}^{C\times i\times D}
\rightarrow\mathbb{R}^{C\times D\times i}$.

Finally, channel mixing is applied after permuting the channel dimension
to the last axis:
\begin{equation}
    \mathcal{M}_i(\mathbf{P}_i)
    =
    \pi_c^{-1}
    \left(
        \pi_c(\mathbf{V}_i)
        +
        \mathcal{C}_i\bigl(\pi_c(\mathbf{V}_i)\bigr)
    \right),
    \qquad
    \mathcal{C}_i:\mathbb{R}^{C}\rightarrow\mathbb{R}^{C},
    \label{eq:hgpm_channel_mixing}
\end{equation}
where
$\pi_c:\mathbb{R}^{C\times i\times D}
\rightarrow\mathbb{R}^{i\times D\times C}$.
Each prefix length uses a separately parameterized
$\mathcal{M}_i$ because its temporal mixer
$\mathcal{T}_i$ must operate on a different number of patch positions.

The schedule neither removes nor merges patches, so the number of
representations remains $N$.
Patch $\mathbf{U}_j$ receives $N-j$ updates: $\mathbf{U}_1$ receives
$N-1$, while $\mathbf{U}_N$ receives none.
Because every processed prefix ends before position $N$,
$\mathbf{U}_N$ is neither updated by nor provided to a prefix mixer.
It reaches the prediction head after only patch projection and
intra-patch refinement.
The head therefore receives progressively mixed historical context
together with an unmixed representation of the interval nearest the
forecast boundary.

\subsection{The CAMP Architecture}
\label{sec:camp}

\paragraph{Cycle Removal and Reinjection.}
The gated historical cycle is subtracted from the normalized input.
The backbone predicts the residual, after which the gated future cycle is
added:
\begin{equation}
\begin{aligned}
    \mathbf{R}
    &=
    \mathbf{X}^{\mathrm{in}}
    -
    \mathbf{C}^{\mathrm{hist}},\\
    \hat{\mathbf{Y}}^{\mathrm{norm}}
    &=
    \hat{\mathbf{R}}
    +
    \mathbf{C}^{\mathrm{pred}}.
\end{aligned}
\label{eq:cycle_residual}
\end{equation}
Here,
$\mathbf{R}\in\mathbb{R}^{C\times L}$ and
$\hat{\mathbf{R}}\in\mathbb{R}^{C\times H}$.
The separate gates control how much cyclic structure is removed from the
input and reintroduced into the forecast.

\paragraph{Multi-Resolution Residual Decomposition.}
CAMP decomposes the residual using a fixed, nonlearnable stationary
wavelet transform:
\begin{equation}
\begin{aligned}
    \operatorname{SWT}(\mathbf{R})
    =
    [
    \mathbf{A}_J,
    \mathbf{D}_J,
    \mathbf{D}_{J-1},
    \ldots,
    \mathbf{D}_1
    ].
\end{aligned}
\label{eq:swt}
\end{equation}
Each stream belongs to $\mathbb{R}^{C\times L}$.
Here, $\mathbf{A}_J$ is the level-$J$ approximation and
$\mathbf{D}_j$ the detail stream at level $j$.

The decomposition uses fixed wavelet analysis filters, dilated grouped
one-dimensional convolutions, and circular padding.
Convolutions are applied independently to each variable, with dilation
doubling at each level.
Because SWT is undecimated, all streams retain length $L$.
Patch position $n$ therefore corresponds to the same temporal interval
across streams, allowing a common patch length and stride.

No inverse SWT is used; the coefficients serve directly as aligned
multi-resolution feature streams.

\paragraph{Stream-Specific Encoding.}
Each of the $J+1$ streams is independently patched, projected, and
processed by its own Intra-Patch Mixer.
The streams are routed as follows:
\begin{itemize}
    \item the approximation stream $\mathbf{A}_J$ is processed by a
    Simple Patch Mixer;
    \item the coarsest detail stream $\mathbf{D}_J$ is processed by
    HGPM;
    \item the finer detail streams
    $\mathbf{D}_{J-1},\ldots,\mathbf{D}_1$ are processed by independent
    Simple Patch Mixers.
\end{itemize}

Only $\mathbf{D}_J$ is routed through HGPM because its coarse residual
oscillations are expected to benefit most from progressive long-range
interaction.
The approximation and finer detail streams use lighter encoders.
This routing is an architectural choice and does not assume that one
wavelet stream is intrinsically more important than the others.

\paragraph{Simple Patch Mixer.}
The Simple Patch Mixer contains three separate MLPs. The feature-wise MLP
mixes the embedding dimension $D$, the temporal-wise MLP mixes the $N$
patches, and the channel-wise MLP mixes the $C$ variables. Each MLP is
applied with a residual connection. The mixer is applied once to the full
patch sequence, and each wavelet stream uses independent encoder parameters.

\paragraph{Aggregation and Prediction.}
Let
$\mathbf{E}_0,\ldots,\mathbf{E}_J
\in\mathbb{R}^{C\times N\times D}$
denote the encoded streams, where $\mathbf{E}_0$ corresponds to
$\mathbf{A}_J$ and $\mathbf{E}_1$ to $\mathbf{D}_J$.
Because their patch positions and representation sizes are aligned, they
are averaged as
$\mathbf{E}=\frac{1}{J+1}\sum_{j=0}^{J}\mathbf{E}_j$.
This aggregation introduces no learnable stream weights and keeps the
prediction-head input size independent of the number of wavelet levels.

For each variable, the patch and feature dimensions are flattened and
mapped to $H$ values by a linear prediction head shared across variables.
This yields
$\hat{\mathbf{R}}\in\mathbb{R}^{C\times H}$.
The future cyclic component is added according to
Equation~\eqref{eq:cycle_residual}, and inverse RevIN returns
$\hat{\mathbf{Y}}\in\mathbb{R}^{C\times H}$ in the original scale.

\section{Experiments}
\subsection{Experimental Setup}

\textbf{Datasets and Baselines.}
We evaluate \textbf{CAMP} on four ETT benchmarks, namely ETTh1, ETTh2, ETTm1, and ETTm2~\cite{informer}, together with Weather, Electricity, and Traffic for long-term forecasting.
These datasets cover diverse domains, temporal resolutions, sequence lengths, and numbers of variables.
For short-term forecasting, we use the PEMS traffic datasets~\cite{pems}.

On the long-term benchmarks, CAMP is compared with SRSNet~\cite{srsnet}, TimeKAN~\cite{timekan}, Amplifier~\cite{amplifier}, iTransformer~\cite{itransformer}, CycleNet~\cite{cyclenet}, PatchTST~\cite{patchtst}, DLinear~\cite{dlinear}, and Crossformer~\cite{crossformer}.
On PEMS, the baselines are CycleNet/MLP~\cite{cyclenet}, RLinear~\cite{dlinear}, iTransformer~\cite{itransformer}, PatchTST~\cite{patchtst}, Crossformer~\cite{crossformer}, DLinear~\cite{dlinear}, and SCINet~\cite{scinet}.

\textbf{Evaluation Protocol.}
For long-term forecasting, we evaluate prediction horizons
$H\in\{96,192,336,720\}$.
Following SRSNet~\cite{srsnet}, the look-back length is selected from
$L\in\{96,336,512\}$, and the best result is reported for every combination of model, dataset, and prediction horizon.
Treating $L$ as a tunable hyperparameter is supported by evidence that the look-back length can substantially affect forecasting performance~\cite{chaindep}.

For PEMS, we fix $L=96$ and evaluate
$H\in\{12,24,48,96\}$.
Data preprocessing and train, validation, and test splits follow previous work~\cite{informer,autoformer}.
For robustness, each CAMP configuration is evaluated using three random seeds, and the mean performance is reported.
CAMP is trained using Mean Squared Error (MSE) loss and evaluated with MSE and Mean Absolute Error (MAE).
Detailed dataset statistics, hyperparameter search spaces, and selected configurations are provided in Supplementary Material~A (Tables~A.1--A.3).

\subsection{Main Results}

\begin{table*}[!t]
    \caption{Average results across forecasting horizons on the standard
    datasets, averaged over three random seeds. The best result for each
    metric is in \textbf{bold}, and the second-best result is
    \underline{underlined}. Win Count is computed over dataset-average rows.
    Because this table reports averages across the four prediction horizons,
    complete horizon-wise results are provided in Supplementary Material~A
    (Table~A.4), and the corresponding random-seed variability is reported in
    Supplementary Material~A (Table~A.5).}
    \label{tab:average-results-standard-datasets}
    \begin{center}
    \scriptsize
    \setlength{\tabcolsep}{2.0pt}
    \renewcommand{\arraystretch}{1}
    \resizebox{\textwidth}{!}{%
    \begin{tabular}{l|cc|cc|cc|cc|cc|cc|cc|cc|cc}
    \specialrule{1pt}{0pt}{0pt}
    Dataset
    & \multicolumn{2}{c|}{Ours}
    & \multicolumn{2}{c|}{SRSNet}
    & \multicolumn{2}{c|}{TimeKAN}
    & \multicolumn{2}{c|}{Amplifier}
    & \multicolumn{2}{c|}{iTransformer}
    & \multicolumn{2}{c|}{CycleNet}
    & \multicolumn{2}{c|}{PatchTST}
    & \multicolumn{2}{c|}{DLinear}
    & \multicolumn{2}{c}{Crossformer} \\
    \midrule
    & MSE & MAE
    & MSE & MAE
    & MSE & MAE
    & MSE & MAE
    & MSE & MAE
    & MSE & MAE
    & MSE & MAE
    & MSE & MAE
    & MSE & MAE \\
    \midrule
  
    ETTm1
    & \textbf{0.343} & \textbf{0.376}
    & 0.351 & \underline{0.378}
    & \underline{0.344} & 0.380
    & 0.353 & 0.379
    & 0.347 & \underline{0.378}
    & 0.361 & 0.390
    & 0.352 & 0.382
    & 0.357 & 0.379
    & 0.431 & 0.443 \\
  
    ETTm2
    & \textbf{0.251} & \textbf{0.313}
    & \underline{0.252} & \underline{0.314}
    & 0.260 & 0.318
    & 0.256 & 0.318
    & 0.258 & 0.318
    & 0.270 & 0.324
    & 0.258 & 0.315
    & 0.267 & 0.332
    & 0.633 & 0.578 \\
  
    ETTh1
    & \textbf{0.397} & \textbf{0.421}
    & \underline{0.404} & \underline{0.424}
    & 0.409 & 0.427
    & 0.421 & 0.433
    & 0.440 & 0.445
    & 0.437 & 0.440
    & 0.417 & 0.431
    & 0.423 & 0.437
    & 0.441 & 0.465 \\
  
    ETTh2
    & \textbf{0.309} & \textbf{0.369}
    & 0.334 & 0.385
    & 0.350 & 0.397
    & 0.356 & 0.402
    & 0.359 & 0.395
    & 0.371 & 0.407
    & \underline{0.331} & \underline{0.379}
    & 0.431 & 0.447
    & 0.835 & 0.676 \\
  
    Electricity
    & \textbf{0.154} & \textbf{0.251}
    & 0.161 & \underline{0.254}
    & 0.164 & 0.258
    & 0.163 & 0.256
    & 0.163 & 0.258
    & \underline{0.157} & \textbf{0.251}
    & 0.162 & \underline{0.254}
    & 0.166 & 0.264
    & 0.293 & 0.351 \\
  
    Weather
    & \textbf{0.219} & \textbf{0.261}
    & 0.226 & 0.266
    & 0.226 & 0.268
    & \underline{0.224} & \underline{0.264}
    & 0.232 & 0.270
    & 0.226 & 0.266
    & 0.230 & 0.265
    & 0.246 & 0.300
    & 0.230 & 0.290 \\
  
    Traffic
    & 0.405 & 0.277
    & \textbf{0.392} & \underline{0.270}
    & 0.420 & 0.285
    & 0.417 & 0.291
    & 0.397 & 0.281
    & 0.413 & 0.281
    & \underline{0.396} & \textbf{0.266}
    & 0.434 & 0.295
    & 0.535 & 0.300 \\
  
    \midrule
    \textbf{Win Count}
    & \textbf{6} & \textbf{6}
    & 1 & 0
    & 0 & 0
    & 0 & 0
    & 0 & 0
    & 0 & 1
    & 0 & 1
    & 0 & 0
    & 0 & 0 \\
    \specialrule{1pt}{0pt}{0pt}
    \end{tabular}%
    }
    \end{center}
  \end{table*}

\begin{table*}[!t]
    \caption{Results over different prediction horizons on PEMS datasets,
    averaged over three random seeds. The best result for each metric is in
    \textbf{bold}, and the second-best result is \underline{underlined}.
    Random-seed variability for CAMP on the PEMS benchmarks is provided in
    Supplementary Material~A (Table~A.7).}
    \label{tab:full-results-pems-datasets}
    \centering
    \scriptsize
    \setlength{\tabcolsep}{1.8pt}
    \renewcommand{\arraystretch}{0.88}

    \resizebox{0.96\textwidth}{!}{%
    \begin{tabular}{l|c|cc|cc|cc|cc|cc|cc|cc|cc|cc}
    \specialrule{1pt}{0pt}{0pt}
    Dataset & Horizon
    & \multicolumn{2}{c|}{Ours}
    & \multicolumn{2}{c|}{CycleNet/MLP}
    & \multicolumn{2}{c|}{CycleNet/Linear}
    & \multicolumn{2}{c|}{RLinear}
    & \multicolumn{2}{c|}{iTransformer}
    & \multicolumn{2}{c|}{PatchTST}
    & \multicolumn{2}{c|}{Crossformer}
    & \multicolumn{2}{c|}{DLinear}
    & \multicolumn{2}{c}{SCINet} \\
    \midrule
    &
    & MSE & MAE
    & MSE & MAE
    & MSE & MAE
    & MSE & MAE
    & MSE & MAE
    & MSE & MAE
    & MSE & MAE
    & MSE & MAE
    & MSE & MAE \\
    \midrule

    PEMS03
    & 12
    & \textbf{0.063} & \textbf{0.168}
    & \underline{0.066} & \underline{0.172}
    & 0.080 & 0.192
    & 0.126 & 0.236
    & 0.071 & 0.174
    & 0.099 & 0.216
    & 0.090 & 0.203
    & 0.122 & 0.243
    & \underline{0.066} & \underline{0.172} \\

    & 24
    & \textbf{0.084} & \textbf{0.194}
    & 0.089 & 0.201
    & 0.120 & 0.237
    & 0.246 & 0.334
    & 0.093 & 0.201
    & 0.142 & 0.259
    & 0.121 & 0.240
    & 0.201 & 0.317
    & \underline{0.085} & \underline{0.198} \\

    & 48
    & 0.132 & 0.248
    & 0.136 & 0.247
    & 0.156 & 0.258
    & 0.551 & 0.529
    & \textbf{0.125} & \textbf{0.236}
    & 0.211 & 0.319
    & 0.202 & 0.317
    & 0.333 & 0.425
    & \underline{0.127} & \underline{0.238} \\

    & 96
    & 0.187 & 0.302
    & 0.182 & \underline{0.282}
    & 0.199 & 0.292
    & 1.057 & 0.787
    & \textbf{0.164} & \textbf{0.275}
    & 0.269 & 0.370
    & 0.262 & 0.367
    & 0.457 & 0.515
    & \underline{0.178} & 0.287 \\
    \cline{1-20}

    PEMS04
    & 12
    & \textbf{0.069} & \textbf{0.173}
    & 0.078 & 0.186
    & 0.089 & 0.201
    & 0.138 & 0.252
    & 0.078 & 0.183
    & 0.105 & 0.224
    & 0.098 & 0.218
    & 0.148 & 0.272
    & \underline{0.073} & \underline{0.177} \\

    & 24
    & \underline{0.085} & \underline{0.195}
    & 0.099 & 0.212
    & 0.127 & 0.245
    & 0.258 & 0.348
    & 0.095 & 0.205
    & 0.153 & 0.275
    & 0.131 & 0.256
    & 0.224 & 0.340
    & \textbf{0.084} & \textbf{0.193} \\

    & 48
    & \underline{0.112} & \underline{0.228}
    & 0.133 & 0.248
    & 0.169 & 0.286
    & 0.572 & 0.544
    & 0.120 & 0.233
    & 0.229 & 0.339
    & 0.205 & 0.326
    & 0.355 & 0.437
    & \textbf{0.099} & \textbf{0.211} \\

    & 96
    & \underline{0.150} & 0.267
    & 0.167 & 0.281
    & 0.189 & 0.293
    & 1.137 & 0.820
    & \underline{0.150} & \underline{0.262}
    & 0.291 & 0.389
    & 0.402 & 0.457
    & 0.452 & 0.504
    & \textbf{0.114} & \textbf{0.227} \\
    \cline{1-20}

    PEMS07
    & 12
    & \textbf{0.057} & \textbf{0.155}
    & \underline{0.062} & \underline{0.162}
    & 0.075 & 0.183
    & 0.118 & 0.235
    & 0.067 & 0.165
    & 0.095 & 0.207
    & 0.094 & 0.200
    & 0.115 & 0.242
    & 0.068 & 0.171 \\

    & 24
    & \textbf{0.074} & \textbf{0.179}
    & \underline{0.086} & 0.192
    & 0.113 & 0.225
    & 0.242 & 0.341
    & 0.088 & \underline{0.190}
    & 0.150 & 0.262
    & 0.139 & 0.247
    & 0.210 & 0.329
    & 0.119 & 0.225 \\

    & 48
    & \textbf{0.105} & \underline{0.218}
    & 0.128 & 0.234
    & 0.157 & 0.254
    & 0.562 & 0.541
    & \underline{0.110} & \textbf{0.215}
    & 0.253 & 0.340
    & 0.311 & 0.369
    & 0.398 & 0.458
    & 0.149 & 0.237 \\

    & 96
    & 0.156 & 0.272
    & 0.176 & 0.268
    & 0.207 & 0.291
    & 1.096 & 0.795
    & \textbf{0.139} & \underline{0.245}
    & 0.346 & 0.404
    & 0.396 & 0.442
    & 0.594 & 0.553
    & \underline{0.141} & \textbf{0.234} \\
    \cline{1-20}

    PEMS08
    & 12
    & \textbf{0.067} & \textbf{0.172}
    & 0.082 & 0.185
    & 0.091 & 0.201
    & 0.133 & 0.247
    & \underline{0.079} & \underline{0.182}
    & 0.168 & 0.232
    & 0.165 & 0.214
    & 0.154 & 0.276
    & 0.087 & 0.184 \\

    & 24
    & \textbf{0.088} & \textbf{0.198}
    & 0.117 & 0.226
    & 0.140 & 0.251
    & 0.249 & 0.343
    & \underline{0.115} & \underline{0.219}
    & 0.224 & 0.281
    & 0.215 & 0.260
    & 0.248 & 0.353
    & 0.122 & 0.221 \\

    & 48
    & \textbf{0.129} & \underline{0.249}
    & \underline{0.169} & 0.268
    & 0.200 & 0.291
    & 0.569 & 0.544
    & 0.186 & \textbf{0.235}
    & 0.321 & 0.354
    & 0.315 & 0.355
    & 0.440 & 0.470
    & 0.189 & 0.270 \\

    & 96
    & \textbf{0.205} & 0.321
    & 0.233 & 0.306
    & 0.272 & 0.328
    & 1.166 & 0.814
    & \underline{0.221} & \textbf{0.267}
    & 0.408 & 0.417
    & 0.377 & 0.397
    & 0.674 & 0.565
    & 0.236 & \underline{0.300} \\

    \midrule
    \textbf{Win Count}
    &
    & \textbf{10} & \textbf{7}
    & 0 & 0
    & 0 & 0
    & 0 & 0
    & \underline{3} & \underline{5}
    & 0 & 0
    & 0 & 0
    & 0 & 0
    & \underline{3} & 4 \\
    \specialrule{1pt}{0pt}{0pt}
    \end{tabular}%
    }
\end{table*}

\begin{table*}[!t]
    \caption{Average ablation results over prediction horizons
$\{96,192,336,720\}$. Each variant is independently hyperparameter-tuned.
``CAMP w/ HGPM Only'' retains only the HGPM-encoded residual branch, while
``CAMP w/ RecurrentCycle'' replaces ACL with the recurrent-cycle module from CycleNet.
The best result for each metric is in \textbf{bold}, and the second-best result
is \underline{underlined}. Horizon-wise ablation results are provided in
Supplementary Material~A (Table~A.6).}
\label{tab:avg-ablation-study}
    \begin{center}
    \scriptsize
    \setlength{\tabcolsep}{5pt}
    \renewcommand{\arraystretch}{1.2}
    \begin{tabular}{l|cc|cc|cc|cc|cc|cc|cc}
    \specialrule{1pt}{0pt}{0pt}
    Model Variant
    & \multicolumn{2}{c|}{ETTm1}
    & \multicolumn{2}{c|}{ETTm2}
    & \multicolumn{2}{c|}{ETTh1}
    & \multicolumn{2}{c|}{ETTh2}
    & \multicolumn{2}{c|}{ECL}
    & \multicolumn{2}{c|}{Weather}
    & \multicolumn{2}{c}{Traffic} \\
    \midrule
    & MSE & MAE
    & MSE & MAE
    & MSE & MAE
    & MSE & MAE
    & MSE & MAE
    & MSE & MAE
    & MSE & MAE \\
    \midrule
    
    w/o ACL
    & 0.352 & 0.384
    & \underline{0.256} & \underline{0.316}
    & \underline{0.406} & 0.424
    & 0.320 & 0.381
    & 0.163 & 0.262
    & 0.227 & 0.266
    & \underline{0.425} & \underline{0.303} \\
    \cline{1-15}
    
    w/o Intra-patch Mixer
    & 0.354 & 0.380
    & \underline{0.256} & \underline{0.316}
    & 0.407 & 0.425
    & 0.319 & 0.378
    & 0.162 & 0.262
    & 0.227 & 0.265
    & 0.432 & 0.310 \\
    \cline{1-15}
    
    w/o SWT
    & \underline{0.350} & 0.381
    & 0.257 & 0.317
    & \underline{0.406} & 0.424
    & 0.328 & 0.382
    & 0.168 & 0.266
    & 0.224 & 0.264
    & 0.432 & 0.308 \\
    \cline{1-15}
    
    w/o HGPM
    & \underline{0.350} & 0.380
    & 0.260 & 0.318
    & 0.410 & 0.426
    & 0.323 & 0.380
    & 0.162 & 0.264
    & 0.226 & 0.265
    & 0.433 & 0.312 \\
    \cline{1-15}

    CAMP w/ HGPM Only
    & 0.359 & 0.382
    & 0.258 & 0.317
    & 0.446 & 0.432
    & 0.343 & 0.393
    & 0.175 & 0.271
    & \underline{0.222} & \underline{0.262}
    & 0.429 & \underline{0.303} \\
    \cline{1-15}
    
    CAMP w/ RecurrentCycle
    & 0.352 & \underline{0.378}
    & 0.258 & 0.317
    & \underline{0.406} & \underline{0.423}
    & \underline{0.318} & \underline{0.377}
    & \underline{0.161} & \underline{0.261}
    & 0.225 & 0.265
    & 0.432 & 0.307 \\
    \cline{1-15}
    
    CAMP
    & \textbf{0.343} & \textbf{0.376}
    & \textbf{0.251} & \textbf{0.313}
    & \textbf{0.397} & \textbf{0.421}
    & \textbf{0.309} & \textbf{0.369}
    & \textbf{0.154} & \textbf{0.251}
    & \textbf{0.219} & \textbf{0.261}
    & \textbf{0.405} & \textbf{0.277} \\
    
    \specialrule{1pt}{0pt}{0pt}
    \end{tabular}
    \end{center}
\end{table*}

\begin{table}[!t]
    \centering
    \caption{Average forecasting performance over prediction horizons
    $\{96,192,336,720\}$ with look-back length $L=336$.
    CycleNet denotes the MLP variant with look-back length $L=336$.
    Imp. reports the relative error reduction achieved by CycleNet+ACL.}
    \label{tab:cyclenet-acl-average}

    \scriptsize
    \setlength{\tabcolsep}{5pt}
    \renewcommand{\arraystretch}{1}

    \resizebox{\columnwidth}{!}{%
    \begin{tabular}{l|cc|cc|cc}
        \toprule
        Dataset
        & \multicolumn{2}{c|}{CycleNet}
        & \multicolumn{2}{c|}{CycleNet+ACL}
        & \multicolumn{2}{c}{Imp. (\%)} \\
        & MSE & MAE & MSE & MAE & MSE & MAE \\
        \midrule

        ETTm1
        & \underline{0.361}
        & \underline{0.390}
        & \textbf{0.352}
        & \textbf{0.379}
        & \textbf{+2.41}
        & \textbf{+2.90} \\

        Traffic
        & \underline{0.413}
        & \underline{0.281}
        & \textbf{0.403}
        & \textbf{0.280}
        & \textbf{+2.40}
        & \textbf{+0.46} \\

        Weather
        & \underline{0.226}
        & \underline{0.266}
        & \textbf{0.223}
        & \textbf{0.260}
        & \textbf{+1.28}
        & \textbf{+2.11} \\

        \bottomrule
    \end{tabular}%
    }
\end{table}

Table~\ref{tab:average-results-standard-datasets} reports the
average performance over the four long-term forecasting horizons.
CAMP obtains the best MSE on six of the seven datasets and the
best or tied-best MAE on six, suggesting generally consistent
performance across datasets with different temporal resolutions
and channel dimensions.

The results show the largest difference on ETTh2, where CAMP achieves
an MSE of $0.309$ and an MAE of $0.369$, compared with $0.331$
and $0.379$ for the second-best PatchTST. CAMP also records the
best results on Weather with $0.219$ MSE and $0.261$ MAE, and
the best MSE on Electricity with $0.154$, while tying CycleNet
for the best MAE of $0.251$. These results suggest that adaptive
cycle learning and multi-resolution residual modeling may be
effective across datasets with different seasonal and frequency
characteristics.

The selected configurations show that CAMP uses a look-back length of
$L=336$ more frequently than $L=512$ across the long-term
forecasting settings. In contrast, the official SRSNet
configurations predominantly adopt $L=512$%
\footnote{\url{https://github.com/decisionintelligence/SRSNet}}.
Since both models search over the same candidate lengths
$L\in\{96,336,512\}$, this pattern suggests that CAMP can often
achieve competitive or lower forecasting errors without relying
on the longest available historical context.

Traffic is the only dataset on which CAMP does not obtain the
best result. SRSNet achieves the lowest MSE of $0.392$, while
PatchTST obtains the lowest MAE of $0.266$, compared with
CAMP's $0.405$ MSE and $0.277$ MAE. This may be related to the
large number of strongly correlated sensor variables in Traffic,
where dense all-to-all channel mixing could introduce irrelevant
dependencies. Prior work suggests that indiscriminate
channel-dependent modeling may weaken channel-specific
representations and cause oversmoothing~\cite{channelmixing}.
More selective channel interaction may therefore be beneficial
for such high-dimensional datasets. CAMP remains competitive with most
baselines on both metrics.

\subsection{Short-Term Forecasting Results on PEMS}

Table~\ref{tab:full-results-pems-datasets} reports results on four PEMS
datasets. Overall, \textbf{CAMP} achieves the highest win count, with the
best MSE in 10 of the 16 settings and the best MAE in 7.

CAMP appears particularly effective at short horizons. At $H=12$, it
achieves the best MSE and MAE on all four datasets, including
$0.063/0.168$ on PEMS03 and $0.057/0.155$ on PEMS07. It also obtains
the best MSE across all four horizons on PEMS08, suggesting strong
performance in capturing short-term traffic variations.

At longer horizons, iTransformer and SCINet perform better in several
settings, particularly in MAE. This may indicate that long-range traffic
forecasting benefits from stronger spatial or cross-sensor modeling.
CAMP remains competitive at medium and long horizons and shows strong
overall performance across the PEMS benchmarks.
\subsection{Ablation Study}

  Table~\ref{tab:avg-ablation-study} reports the average results over prediction horizons $\{96,192,336,720\}$. We compare CAMP with six independently tuned variants. \textit{w/o ACL}, \textit{w/o Intra-patch Mixer}, \textit{w/o SWT}, and \textit{w/o HGPM} remove the corresponding modules. \textit{CAMP w/ HGPM Only} retains only the HGPM-encoded residual branch instead of aggregating all SWT streams, while \textit{CAMP w/ RecurrentCycle} replaces ACL with a recurrent-cycle module.
  
  CAMP achieves the best MSE and MAE on all seven datasets, showing that its components provide complementary benefits. Removing ACL consistently reduces accuracy, and replacing it with RecurrentCycle also performs worse, supporting the use of adaptive frequency-based cycle learning. The performance drops without SWT or when using only the HGPM branch further demonstrate the importance of combining aligned multi-resolution streams.
  
  Removing either the intra-patch mixer or HGPM also degrades performance, indicating that local modeling within individual patches and progressive information aggregation across patches are both beneficial. Overall, the results show that CAMP's performance arises from the joint use of adaptive cycle learning, multi-resolution residual decomposition, and complementary patch-mixing operations.
\subsection{Generalizability of Adaptive Cycle Learning}

To evaluate whether ACL is beneficial beyond CAMP, we integrate it into CycleNet and report the average results in Table~\ref{tab:cyclenet-acl-average}. For each dataset, CycleNet denotes the MLP variant with look-back length $L=336$. Adding ACL improves both MSE and MAE on all three datasets, demonstrating that adaptive frequency identification can also enhance an existing cycle-based forecasting model. The largest improvements are observed on Traffic in MSE and Weather in MAE, with relative error reductions of 2.40\% and 2.11\%, respectively. These results suggest that ACL can serve as a general cycle-learning component rather than being effective only within CAMP.

\section{Conclusion and Future Work}

We introduced CAMP, a cycle-aware forecasting framework that combines
sample-specific cycle identification, multi-resolution residual modeling,
and horizon-guided patch mixing. By separating cyclic and residual
dynamics, CAMP appears to capture complementary temporal structures and
achieves strong performance across both long- and short-term benchmarks.
The ablation results further suggest that each major component contributes
to the overall performance.

CAMP has higher computational cost, while FFT-based
cycle estimation may be less reliable for short or weakly periodic inputs.
Dense channel interaction may also be less suitable for highly
multivariate datasets. Future work will therefore explore more efficient
patch mixing, more robust cycle estimation, and selective channel-aware
modeling.
\bibliographystyle{aaai2027}
\bibliography{aaai2027}

@misc{weather,
  author = {{Max Planck Institute for Biogeochemistry}},
  title = {Jena Climate Dataset},
  year = {2024},
  howpublished = {\url{https://www.bgc-jena.mpg.de/wetter/}},
  note = {Accessed: 2024-05-20}
}

@misc{electricity,
  author       = {Trindade, Artur},
  title        = {{ElectricityLoadDiagrams20112014}},
  year         = {2015},
  howpublished = {UCI Machine Learning Repository},
  note         = {{DOI}: https://doi.org/10.24432/C58C86}
}

@misc{pems,
  author = {{California Department of Transportation (Caltrans)}},
  title = {Performance Measurement System ({PeMS})},
  year = {2024},
  howpublished = {\url{https://pems.dot.ca.gov/}},
  note = {Accessed: 2024-05-20}
}

@inproceedings{patchtst,
title={A Time Series is Worth 64 Words:  Long-term Forecasting with Transformers},
author={Yuqi Nie and Nam H Nguyen and Phanwadee Sinthong and Jayant Kalagnanam},
booktitle={International Conference on Learning Representations},
year={2023}
}

@inproceedings{
itransformer,
title={iTransformer: Inverted Transformers Are Effective for Time Series Forecasting},
author={Yong Liu and Tengge Hu and Haoran Zhang and Haixu Wu and Shiyu Wang and Lintao Ma and Mingsheng Long},
booktitle={International Conference on Learning Representations},
year={2024}
}

@inproceedings{
simpletm,
title={Simple{TM}: A Simple Baseline for Multivariate Time Series Forecasting},
author={Hui Chen and Viet Luong and Lopamudra Mukherjee and Vikas Singh},
booktitle={International Conference on Learning Representations},
year={2025}
}

@inproceedings{timemixer,
	title={TimeMixer: Decomposable Multiscale Mixing for Time Series Forecasting},
	author={Wang, Shiyu and Wu, Haixu and Shi, Xiaoming and Hu, Tengge and Luo, Huakun and Ma, Lintao and Zhang, James Y and ZHOU, JUN},
	booktitle={International Conference on Learning Representations},
	year={2024}
}

@article{tsmixer,
title={{TSM}ixer: An All-{MLP} Architecture for Time Series Forecasting},
author={Si-An Chen and Chun-Liang Li and Sercan O Arik and Nathanael Christian Yoder and Tomas Pfister},
journal={Transactions on Machine Learning Research},
year={2023}
}

@article{waveform, 
title={WaveForM: Graph Enhanced Wavelet Learning for Long Sequence Forecasting of Multivariate Time Series}, 
abstractNote={Multivariate time series (MTS) analysis and forecasting are crucial in many real-world applications, such as smart traffic management and weather forecasting. However, most existing work either focuses on short sequence forecasting or makes predictions predominantly with time domain features, which is not effective at removing noises with irregular frequencies in MTS. Therefore, we propose WaveForM, an end-to-end graph enhanced Wavelet learning framework for long sequence FORecasting of MTS. WaveForM first utilizes Discrete Wavelet Transform (DWT) to represent MTS in the wavelet domain, which captures both frequency and time domain features with a sound theoretical basis. To enable the effective learning in the wavelet domain, we further propose a graph constructor, which learns a global graph to represent the relationships between MTS variables, and graph-enhanced prediction modules, which utilize dilated convolution and graph convolution to capture the correlations between time series and predict the wavelet coefficients at different levels. Extensive experiments on five real-world forecasting datasets show that our model can achieve considerable performance improvement over different prediction lengths against the most competitive baseline of each dataset.}, 
journal={Proceedings of the AAAI Conference on Artificial Intelligence}, 
author={Yang, Fuhao and Li, Xin and Wang, Min and Zang, Hongyu and Pang, Wei and Wang, Mingzhong}, 
year={2023}, 
month={Jun.}}

@misc{cyclenet,
      title={CycleNet: Enhancing Time Series Forecasting through Modeling Periodic Patterns}, 
      author={Shengsheng Lin and Weiwei Lin and Xinyi Hu and Wentai Wu and Ruichao Mo and Haocheng Zhong},
      year={2024},
      eprint={2409.18479},
      archivePrefix={arXiv},
      primaryClass={cs.LG},
      url={https://arxiv.org/abs/2409.18479}, 
}

@Inbook{swt,
author="Nason, G. P.
and Silverman, B. W.",

title="The Stationary Wavelet Transform and some Statistical Applications",
bookTitle="Wavelets and Statistics",
year="1995",
publisher="Springer New York",
address="New York, NY",
pages="281--299",
isbn="978-1-4612-2544-7"
}

@inproceedings{dlinear,
  title={Are transformers effective for time series forecasting?},
  author={Zeng, Ailing and Chen, Muxi and Zhang, Lei and Xu, Qiang},
  booktitle={Proceedings of the AAAI Conference on Artificial Intelligence},
  year={2023}
}

@article{hdmixer, title={HDMixer: Hierarchical Dependency with Extendable Patch for Multivariate Time Series Forecasting}, volume={38}, url={https://ojs.aaai.org/index.php/AAAI/article/view/29155}, DOI={10.1609/aaai.v38i11.29155}, abstractNote={Multivariate time series (MTS) prediction has been widely adopted in various scenarios. Recently, some methods have employed patching to enhance local semantics and improve model performance. However, length-fixed patch are prone to losing temporal boundary information, such as complete peaks and periods. Moreover, existing methods mainly focus on modeling long-term dependencies across patches, while paying little attention to other dimensions (e.g., short-term dependencies within patches and complex interactions among cross-variavle patches). To address these challenges, we propose a pure MLP-based HDMixer, aiming to acquire patches with richer semantic information and efficiently modeling hierarchical interactions. Specifically, we design a Length-Extendable Patcher (LEP) tailored to MTS, which enriches the boundary information of patches and alleviates semantic incoherence in series. Subsequently, we devise a Hierarchical Dependency Explorer (HDE) based on pure MLPs. This explorer effectively models short-term dependencies within patches, long-term dependencies across patches, and complex interactions among variables. Extensive experiments on 9 real-world datasets demonstrate the superiority of our approach. The code is available at https://github.com/hqh0728/HDMixer.}, journal={Proceedings of the AAAI Conference on Artificial Intelligence}, author={Huang, Qihe and Shen, Lei and Zhang, Ruixin and Cheng, Jiahuan and Ding, Shouhong and Zhou, Zhengyang and Wang, Yang}, year={2024}, month={Mar.}, pages={12608–12616} }

@misc{wpmixer,
      title={WPMixer: Efficient Multi-Resolution Mixing for Long-Term Time Series Forecasting}, 
      author={Md Mahmuddun Nabi Murad and Mehmet Aktukmak and Yasin Yilmaz},
      year={2024},
      eprint={2412.17176},
      archivePrefix={arXiv},
      primaryClass={cs.LG},
      url={https://arxiv.org/abs/2412.17176}, 
}

@inproceedings{tsmixer2,
  title={Tsmixer: Lightweight mlp-mixer model for multivariate time series forecasting},
  author={Ekambaram, Vijay and Jati, Arindam and Nguyen, Nam and Sinthong, Phanwadee and Kalagnanam, Jayant},
  booktitle={Proceedings of the 29th ACM SIGKDD Conference on Knowledge Discovery and Data Mining},
  year={2023}
}

@inproceedings{vit,
title={An Image is Worth 16x16 Words: Transformers for Image Recognition at Scale},
author={Alexey Dosovitskiy and Lucas Beyer and Alexander Kolesnikov and Dirk Weissenborn and Xiaohua Zhai and Thomas Unterthiner and Mostafa Dehghani and Matthias Minderer and Georg Heigold and Sylvain Gelly and Jakob Uszkoreit and Neil Houlsby},
booktitle={International Conference on Learning Representations},
year={2021}
}

@article{fft,
  title={What is the fast Fourier transform?},
  author={Cochran, William T and Cooley, James W and Favin, David L and Helms, Howard D and Kaenel, Reginald A and Lang, William W and Maling, George C and Nelson, David E and Rader, Charles M and Welch, Peter D},
  journal={Proceedings of the IEEE},
  volume={55},
  number={10},
  pages={1664--1674},
  year={1967},
  publisher={IEEE}
}

@article{frets,
  title={Frequency-domain MLPs are more effective learners in time series forecasting},
  author={Yi, Kun and Zhang, Qi and Fan, Wei and Wang, Shoujin and Wang, Pengyang and He, Hui and An, Ning and Lian, Defu and Cao, Longbing and Niu, Zhendong},
  journal={Advances in Neural Information Processing Systems},
  year={2023}
}

@inproceedings{
crossformer,
title={Crossformer: Transformer Utilizing Cross-Dimension Dependency for Multivariate Time Series Forecasting},
author={Yunhao Zhang and Junchi Yan},
booktitle={International Conference on Learning Representations},
year={2023},
url={https://openreview.net/forum?id=vSVLM2j9eie}
}

@article{autoformer,
  author       = {Haixu Wu and
                  Jiehui Xu and
                  Jianmin Wang and
                  Mingsheng Long},
  title        = {Autoformer: Decomposition Transformers with Auto-Correlation for Long-Term
                  Series Forecasting},
  journal      = {CoRR},
  volume       = {abs/2106.13008},
  year         = {2021},
  url          = {https://arxiv.org/abs/2106.13008},
  eprinttype   = {arXiv},
  eprint       = {2106.13008},
  bibsource    = {dblp computer science bibliography, https://dblp.org}
}

@article{informer,
  author       = {Haoyi Zhou and
                  Shanghang Zhang and
                  Jieqi Peng and
                  Shuai Zhang and
                  Jianxin Li and
                  Hui Xiong and
                  Wancai Zhang},
  title        = {Informer: Beyond Efficient Transformer for Long Sequence Time-Series
                  Forecasting},
  journal      = {CoRR},
  volume       = {abs/2012.07436},
  year         = {2020},
  url          = {https://arxiv.org/abs/2012.07436},
  eprinttype   = {arXiv},
  eprint       = {2012.07436},
  bibsource    = {dblp computer science bibliography, https://dblp.org}
}

@misc{srsnet,
      title={Enhancing Time Series Forecasting through Selective Representation Spaces: A Patch Perspective}, 
      author={Xingjian Wu and Xiangfei Qiu and Hanyin Cheng and Zhengyu Li and Jilin Hu and Chenjuan Guo and Bin Yang},
      year={2025},
      eprint={2510.14510},
      archivePrefix={arXiv},
      primaryClass={cs.LG},
      url={https://arxiv.org/abs/2510.14510}, 
}

@misc{timekan,
      title={TimeKAN: KAN-based Frequency Decomposition Learning Architecture for Long-term Time Series Forecasting}, 
      author={Songtao Huang and Zhen Zhao and Can Li and Lei Bai},
      year={2025},
      eprint={2502.06910},
      archivePrefix={arXiv},
      primaryClass={cs.LG},
      url={https://arxiv.org/abs/2502.06910}, 
}

@misc{amplifier,
      title={Amplifier: Bringing Attention to Neglected Low-Energy Components in Time Series Forecasting}, 
      author={Jingru Fei and Kun Yi and Wei Fan and Qi Zhang and Zhendong Niu},
      year={2025},
      eprint={2501.17216},
      archivePrefix={arXiv},
      primaryClass={cs.LG},
      url={https://arxiv.org/abs/2501.17216}, 
}

@misc{freqcycle,
      title={FreqCycle: A Multi-Scale Time-Frequency Analysis Method for Time Series Forecasting}, 
      author={Boya Zhang and Shuaijie Yin and Huiwen Zhu and Xing He},
      year={2026},
      eprint={2603.09661},
      archivePrefix={arXiv},
      primaryClass={cs.LG},
      url={https://arxiv.org/abs/2603.09661}, 
}

@article{depts,
  title={DEPTS: Deep expansion learning for periodic time series forecasting},
  author={Fan, Wei and Zheng, Shun and Yi, Xiaohan and Cao, Wei and Fu, Yanjie and Bian, Jiang and Liu, Tie-Yan},
  journal={arXiv preprint arXiv:2203.07681},
  year={2022}
}

@article{sparsetsf,
  title={Sparsetsf: Modeling long-term time series forecasting with 1k parameters},
  author={Lin, Shengsheng and Lin, Weiwei and Wu, Wentai and Chen, Haojun and Yang, Junjie},
  journal={arXiv preprint arXiv:2405.00946},
  year={2024}
}

@inproceedings{revin,
  title={Reversible instance normalization for accurate time-series forecasting against distribution shift},
  author={Kim, Taesung and Kim, Jinhee and Tae, Yunwon and Park, Cheonbok and Choi, Jang-Ho and Choo, Jaegul},
  booktitle={International Conference on Learning Representations},
  year={2021}
}

@inproceedings{chaindep,
  title={Channel Dependence, Limited Lookback Windows, and the Simplicity of Datasets: How Biased is Time Series Forecasting?},
  author={Abdelmalak, Ibram and Madhusudhanan, Kiran and Choi, Jungmin and Kl{\"o}tergens, Christian and Yalavarthi, Vijaya Krishna and Stubbemann, Maximilian and Schmidt-Thieme, Lars},
  booktitle={Pacific-Asia Conference on Knowledge Discovery and Data Mining},
  pages={585--597},
  year={2026},
  organization={Springer}
}

@article{scinet,
  title={Scinet: Time series modeling and forecasting with sample convolution and interaction},
  author={Liu, Minhao and Zeng, Ailing and Chen, Muxi and Xu, Zhijian and Lai, Qiuxia and Ma, Lingna and Xu, Qiang},
  journal={Advances in Neural Information Processing Systems},
  volume={35},
  pages={5816--5828},
  year={2022}
}

@misc{timesnet,
      title={TimesNet: Temporal 2D-Variation Modeling for General Time Series Analysis}, 
      author={Haixu Wu and Tengge Hu and Yong Liu and Hang Zhou and Jianmin Wang and Mingsheng Long},
      year={2023},
      eprint={2210.02186},
      archivePrefix={arXiv},
      primaryClass={cs.LG},
      url={https://arxiv.org/abs/2210.02186}, 
}

@misc{koopa,
      title={Koopa: Learning Non-stationary Time Series Dynamics with Koopman Predictors}, 
      author={Yong Liu and Chenyu Li and Jianmin Wang and Mingsheng Long},
      year={2023},
      eprint={2305.18803},
      archivePrefix={arXiv},
      primaryClass={cs.LG},
      url={https://arxiv.org/abs/2305.18803}, 
}

@misc{fedformer,
      title={FEDformer: Frequency Enhanced Decomposed Transformer for Long-term Series Forecasting}, 
      author={Tian Zhou and Ziqing Ma and Qingsong Wen and Xue Wang and Liang Sun and Rong Jin},
      year={2022},
      eprint={2201.12740},
      archivePrefix={arXiv},
      primaryClass={cs.LG},
      url={https://arxiv.org/abs/2201.12740}, 
}

@misc{film,
      title={FiLM: Frequency improved Legendre Memory Model for Long-term Time Series Forecasting}, 
      author={Tian Zhou and Ziqing Ma and Xue wang and Qingsong Wen and Liang Sun and Tao Yao and Wotao Yin and Rong Jin},
      year={2022},
      eprint={2205.08897},
      archivePrefix={arXiv},
      primaryClass={cs.LG},
      url={https://arxiv.org/abs/2205.08897}, 
}

@article{basisformer,
  title={Basisformer: Attention-based time series forecasting with learnable and interpretable basis},
  author={Ni, Zelin and Yu, Hang and Liu, Shizhan and Li, Jianguo and Lin, Weiyao},
  journal={Advances in Neural Information Processing Systems},
  volume={36},
  pages={71222--71241},
  year={2023}
}

@misc{channelmixing,
      title={From Similarity to Superiority: Channel Clustering for Time Series Forecasting}, 
      author={Jialin Chen and Jan Eric Lenssen and Aosong Feng and Weihua Hu and Matthias Fey and Leandros Tassiulas and Jure Leskovec and Rex Ying},
      year={2024},
      eprint={2404.01340},
      archivePrefix={arXiv},
      primaryClass={cs.LG},
      url={https://arxiv.org/abs/2404.01340}, 
}

@misc{fouriergnn,
      title={FourierGNN: Rethinking Multivariate Time Series Forecasting from a Pure Graph Perspective}, 
      author={Kun Yi and Qi Zhang and Wei Fan and Hui He and Liang Hu and Pengyang Wang and Ning An and Longbing Cao and Zhendong Niu},
      year={2023},
      eprint={2311.06190},
      archivePrefix={arXiv},
      primaryClass={cs.LG},
      url={https://arxiv.org/abs/2311.06190}, 
}

\end{document}